\def\year{2019}\relax
\documentclass[letterpaper]{article} 
\usepackage{aaai19}  
\usepackage{times}  
\usepackage{helvet}  
\usepackage{courier}  
\usepackage{url}  
\usepackage{graphicx}  
\usepackage{bm}
\usepackage{algorithm}
\usepackage{algorithmic}
\usepackage{amsmath}
\usepackage{mathtools}
\usepackage{amssymb}
\usepackage{amsthm}
\usepackage{array}
\usepackage{amsfonts}
\usepackage{color}
\usepackage{url}
\usepackage{multirow}
\usepackage{comment}
\usepackage{wrapfig,lipsum,booktabs}
\newcommand{\citet}[1]{\citeauthor{#1}~\shortcite{#1}}
\newcommand{\citep}{\cite}

\begin{document}
%
\title{Non-Autoregressive Neural Machine Translation with Enhanced Decoder Input}

\author{Junliang Guo$^\dag$\thanks{The work was done when the first author was an intern at Microsoft Research Asia.}, Xu Tan$^\ddag$, Di He$^\S$, Tao Qin$^\ddag$, Linli Xu$^\dag$ \and Tie-Yan Liu$^\ddag$ \\
$^\dag$Anhui Province Key Laboratory of Big Data Analysis and Application,\\
School of Computer Science and Technology,
University of Science and Technology of China \\
$^\ddag$Microsoft Research \\
$^\S$Key Laboratory of Machine Perception (MOE), School of EECS, Peking University \\
$^\dag$guojunll@mail.ustc.edu.cn, linlixu@ustc.edu.cn,
$^\ddag$\{xuta,taoqin,tyliu\}@microsoft.com, $^\S$di\_he@pku.edu.cn
}

\maketitle
\begin{abstract}

Non-autoregressive translation (NAT) models, which remove the dependence on previous target tokens from the inputs of the decoder, achieve significantly inference speedup but at the cost of inferior accuracy compared to autoregressive translation (AT) models. Previous work shows that the quality of the inputs of the decoder is important and largely impacts the model accuracy. In this paper, we propose two methods to enhance the decoder inputs so as to improve NAT models. The first one directly leverages a phrase table generated by conventional SMT approaches to translate source tokens to target tokens, which are then fed into the decoder as inputs. The second one transforms source-side word embeddings to target-side word embeddings through sentence-level alignment and word-level adversary learning, and then feeds the transformed word embeddings into the decoder as inputs. Experimental results show our method largely outperforms 
the NAT baseline~\citep{gu2017non}
by $5.11$ BLEU scores on WMT14 English-German task and $4.72$ BLEU scores on WMT16 English-Romanian task.

\end{abstract}

\section{Introduction}
\label{sec:intro}
The neural network based encoder-decoder framework has achieved very promising performance for machine translation and different network architectures have been proposed, including RNNs ~\citep{sutskever2014sequence,bahdanau2014neural,cho2014properties,wu2016google}, CNNs~\citep{gehring2017convolutional}, and self-attention based Transformer~\citep{vaswani2017attention}. All those models translate a source sentence in an \emph{autoregressive} manner, i.e., they generate a target sentence word by word from left to right~\citep{beyond_error} and the generation of $t$-th token $y_t$ depends on previously generated tokens $y_{1:t-1}$:
\begin{equation}
\label{equ:at_manner}
y_t = \mathbb{D}(y_{1:t-1},\mathbb{E}(x)),
\end{equation}
where $\mathbb{E}(\cdot)$ and $\mathbb{D}(\cdot)$ denote the encoder and decoder of the model respectively, $x$ is the source sentence and $\mathbb{E}(x)$ is the output of the encoder, i.e., the set of hidden representations in the top layer of the encoder.

Since AT models generate target tokens sequentially, the inference speed becomes a bottleneck for real-world translation systems, in which fast response and low latency are expected. To speed up the inference of machine translation, non-autoregressive models~\citep{gu2017non} have been proposed, which generate all target tokens independently and simultaneously.  Instead of using previously generated tokens as in AT models, NAT models take other global signals derived from the source sentence as input. Specifically, Non-AutoRegressive Transformer~(NART)~\citep{gu2017non} takes a copy of source sentence $x$ as the decoder input, and the copy process is guided by fertilities~\citep{brown1993mathematics} which represents how many times each source token will be copied; after that all target tokens are simultaneously predicted:  
\begin{equation}
\label{equ:nat_manner}
y_t = \mathbb{D}(\hat{x},\mathbb{E}(x)),
\end{equation}
where $\hat{x}=(\hat{x}_{1}, ..., \hat{x}_{T_{y}})$ is the copied source sentence and $T_y$ is the length of the target sentence $y$.

  While NAT models significantly reduce the inference latency, they suffer from accuracy degradation compared with their autoregressive counterparts. We notice that the encoder of AT models and that of NAT models are the same; the differences lie in the decoder. In AT models, the generation of the $t$-th token $y_t$ is conditioned on previously generated tokens $y_{1:t-1}$, which provides strong target side context information. In contrast, as NART models generate tokens in parallel, there is no such target-side information available. Although the fertilities are learned to cover target-side information in NART~\citep{gu2017non}, such information contained in the copied source tokens $\hat{x}$ guided by fertilities is indirect and weak because the copied tokens are still in the domain of source language, while the inputs of the decoder of AT models are target-side tokens $y_{1:t-1}$. Consequently, the decoder of a NAT model has to handle the translation task conditioned on less and weaker information compared with its AT counterpart, thus leading to inferior accuracy. As verified by our study (see Figure~\ref{fig:buckets} and Table~\ref{tab:case_study}), NART performs poorly for long sentences, which need stronger target-side conditional information for correct translation than short sentences.

In this paper, we aim to enhance the decoder inputs of NAT models so as to reduce the difficulty of the task that the decoder needs to handle. Our basic idea is to directly feed target-side tokens as the inputs of the decoder. We propose two concrete methods to generate the decoder input $\hat{y}=(\hat{y}_{1},...,\hat{y}_{T_{y}})$ which contains coarse target-side information. The first one is based on a phrase table, and explicitly translates source tokens into target-side tokens through such a pre-trained phrase table. The second one linearly maps the embeddings of source tokens into the target-side embedding space and then the mapped embeddings are fed into the decoder. The mapping is learned in an end-to-end manner by minimizing the $L_{2}$ distance of the mapped source and target embeddings in the sentence level as well as the adversary loss between the mapped source embeddings and target embeddings in the word level.
 
With target-side information as inputs, the decoder works as follows:
\begin{equation}
\label{equ:nat_manner_ours}
y_t = \mathbb{D}(\hat{y},\mathbb{E}(x)),
\end{equation}
where $\hat{y}$ is the enhanced decoder input provided by our methods. 
The decoder now can generate all $y_t$'s in parallel conditioned on the global information $\hat{y}$, which is more close to the target tokens $y_{1:t-1}$ as in the AT model.  In this way, the difficulty of the task for the decoder is largely reduced. 

We conduct experiments on three tasks to verify the proposed method. On WMT14 English-German, WMT16 English-Romanian and IWSLT14 German-English translation tasks, our model outperforms all compared non-autoregressive baseline models. Specifically, we obtain BLEU scores of $24.28$ and $34.51$ which outperform the non-autoregressive baseline~($19.17$ and $29.79$ reported in \citet{gu2017non}) on WMT14 En-De and WMT16 En-Ro tasks.

\section{Background}
\label{sec:pre}

\subsection{Autoregressive Neural Machine Translation}
\label{sec:pre_at}
Deep neural network with encoder-decoder framework has achieved great success on machine translation, with different choices of architectures such as recurrent neural networks~(RNNs)~\citep{bahdanau2014neural,cho2014learning}, convolutional neural networks~(CNNs)~\citep{gehring2017convolutional}, as well as self-attention based transformer~\citep{vaswani2017attention,layerwise}. Early RNNs based models have an inherently sequential architecture that prevents them from being 
parallelized
during training and inference, which is partially solved by CNNs and self-attention based models~\citep{kalchbrenner2016neural,gehring2017convolutional,shen2018dense,vaswani2017attention,song2018double}. Since the entire target translation is exposed to the model at training time, each input token of the decoder is the previous ground truth token and the whole training can be parallel given the well-designed CNNs or self-attention models. However, the autoregressive nature still creates a bottleneck at inference stage, since without ground truth, the prediction of each target token has to condition on previously predicted tokens. See Table~\ref{tab:comparison_models} for a clear comparison between models about whether they are parallelizable.
\begin{table}[htb]
\small
\centering
\begin{tabular}{l|c|cc}
\toprule
\multicolumn{2}{c|}{Models} & Training & Inference \\
\midrule
\multirow{3}{*}{AT models} & RNNs based & $\times$ & $\times$ \\
& CNNs based & $\surd$ & $\times$ \\
& Self-Attention based & $\surd$ & $\times$ \\
\midrule
\multicolumn{2}{c|}{NAT models} & $\surd$ & $\surd$ \\
\bottomrule
\end{tabular}
\caption{Comparison between Autoregressive Translation~(AT) and Non-Autoregressive Translation~(NAT) models about whether they are parallelizable in different stages.}
\label{tab:comparison_models}
\end{table}

\subsection{Non-Autoregressive Neural Machine Translation}
\label{sec:pre_nat}
We generally denote the decoder input as $z=(z_{1}, ...,z_{T_{y}})$ to be consistent in the rest of our paper, which represents $\hat{x}$ and $\hat{y}$ in Equation~(\ref{equ:nat_manner}) and~(\ref{equ:nat_manner_ours}).
The recently proposed non-autoregressive model NART~\citep{gu2017non} breaks the inference bottleneck by exposing all decoder inputs to the network simultaneously. 
The generation of $z$ is guided by 
the fertility prediction function which represents how many target tokens that each source token can translate to, and then repeatedly copy source tokens w.r.t their corresponding fertilities as the decoder input $z$. 
Given $z$, the conditional probability $P(y|x)$ is defined as:
\begin{equation} \small
\begin{aligned}
P(y|x,z)&=\prod_{t=1}^{T_{y}}P(y_{t}|z,x)=\prod_{t=1}^{T_{y}}P(y_{t}|z,\mathbb{E}(x;\theta_{\textrm{enc}});\theta_{\textrm{dec}}), \\
\end{aligned}
\end{equation}
where $T_{y}$ is the length of target sentence, which equals to the summation of all fertility numbers. $\theta_{\textrm{enc}}$ and $\theta_{\textrm{dec}}$ denote the parameter of the encoder and decoder. The negative log-likelihood loss function for NAT model becomes:
\begin{equation} \small
\label{equ:neg_loss}
L_{\textrm{neg}}(x,y;\theta_{\textrm{enc}}, \theta_{\textrm{dec}}) = -\sum_{t=1}^{T_{y}} \log P(y_{t}|z,x)
\end{equation}

\begin{figure*}[tb]
\centering
\centerline{\includegraphics[width=1.6\columnwidth]{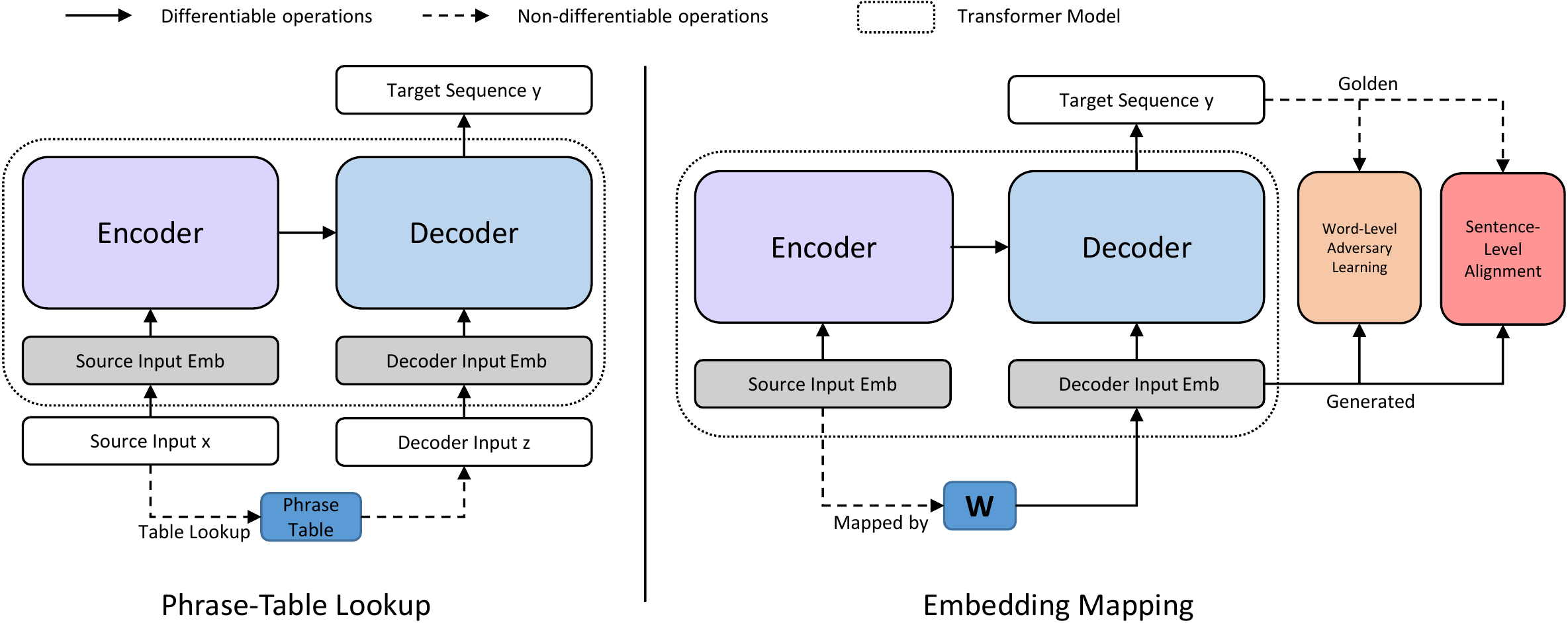}}
\caption{The architecture of our model. A concrete description of fine-grained modules can be found in Section~\ref{sec:arch}.}
\label{fig:model}
\end{figure*}

Although non-autoregressive models can achieve $15\times$ speedup compared to autoregressive models, they are also suffering from accuracy degradation.
Since the conditional dependencies within the target sentence ($y_t$ depends on $y_{1:t-1}$) are removed from the decoder input, the decoder is unable to leverage the inherent sentence structure for prediction.  Hence the decoder has to figure out such target-side information by itself just with the source-side information during training, which is a much more challenging task compared to its autoregressive counterpart. From our study, we find the NART model fails to handle the target sentence generation well. It usually generates repetitive and semantically incoherent sentences with missing words, as shown in Table~\ref{tab:case_study}. Therefore, strong conditional signals should be introduced as the decoder input to help the model learn better internal dependencies within a sentence.

\section{Methodology}
\label{sec:model}

As discussed in Section~\ref{sec:intro}, to improve the accuracy of NAT models, we need to enhance the inputs of the decoder. We introduce our model, Enhanced Non-Autoregressive Transformer~(ENAT), in this section.
We design two kinds of enhanced inputs: one is token level enhancement based on phrase-table lookup and the other one is embedding level enhancement based on embedding mapping. The illustration of the phrase-table lookup and embedding mapping can be found in Figure~\ref{fig:model}.


\subsection{Phrase-Table Lookup}
Previous NAT models take tokens in the source language in as decoder inputs, which make the decoding task difficult. Considering that AT models takes (already generated) target tokens as inputs, a straightforward idea to enhance decoder inputs is to also feed tokens in the target language into the decoder of NAT models. Given a source sentence, a simple method to get target tokens is to translate those source tokens to target tokens using a phrase table,  which brings negligible latency in inference.

To implement this idea, we pre-train a phrase table based on the bilingual training corpus utilizing Moses~\citep{koehn2007moses}, an open-source statistic machine translation~(SMT) toolkit. We then greedily segment the source sentence into $T_{p}$ phrases and translate the phrases one by one according to the phrase table. The details are as follows.  We first calculate the maximum length $L$ among all the phrases contained in the phrase table. For $i$-th source word $x_{i}$, we first check whether phrase $x_{i:i+L}$ has a translation in the phrase table; if not then check $x_{i:i+L-1}$, and so on. If there exists a phrase translation for $x_{i:i+L-j}$, then translate it and check the translation started at $x_{i+L-j+1}$ following the same strategy. This procedure only brings $0.14$ms latency per sentence on average over the \texttt{newstest2014} test set on an Intel Xeon E5-2690 CPU, which is negligible compared with the whole inference latency (e.g., $25$ to $200$+ ms) of the NAT model, as shown in Table~\ref{tab:bleu_results}. 

Note that to reduce inference latency, we only search the phrase table to obtain a course phrase-to-phrase translation, without utilizing the full procedure~(including language model scoring and tree-based searching). During inference, we generate $z$ by the phrase table lookup and skip phrases that do not have translations.

\subsection{Embedding Mapping}
As the phrase table is pre-trained from SMT systems, it cannot be updated/optimized during NAT model training, and may lead to poor translation quality if the table is not very accurate. Therefore, we propose the embedding mapping approach, which first linearly maps the source token embeddings to target embeddings and feeds them into the decoder as inputs. This linear mapping can be trained end-to-end together with NAT models.

To be concrete, given the source sentence $x=(x_{1},...,x_{T_{x}})$ and its corresponding embedding matrix $E_{x} \in \mathbb{R}^{T_{x} \times d}$ where $d$ is the dimensionality of embeddings, we transform $E_{x}$ into the target-side embedding space by a linear mapping function $f_G$:
\begin{equation} \small
\label{equ:transform_ex}
E_{\tilde{z}} = f_{G}(E_{x}; W)=E_{x} W,
\end{equation}
where $W \in \mathbb{R}^{d \times d}$ is the projection matrix to be learned and $E_{\tilde{z}} \in \mathbb{R}^{T_{x} \times d}$ is the decoder input candidate who has the same number of tokens as the source sentence $x$. We then reconstruct $E_{\tilde{z}} \in \mathbb{R}^{T_{x} \times d}$ to the final decoder input $E_{z} \in \mathbb{R}^{T_{y} \times d}$ whose length is identical to the length of target sentence by a simple method which will be introduced in the next section. Intuitively, $E_{z}$ should contain coarse target-side information, which is the translation of the corresponding source tokens in the embedding space, although in similar order as the source tokens. To ensure the projection matrix $W$ to be learned end-to-end with the NAT model, we regularize the learning of $W$ with sentence-level alignment and word-level adversary learning. 

Since we already have the sentence-level alignment from the training set, we can minimize the $L_{2}$ distance between the mapped source embeddings and the ground truth target embeddings in the sentence level:
\begin{equation}
\label{equ:align}
L_{\textrm{align}}(x,y) = \Vert f_{G}(e(x)) - e(y) \Vert_{2},
\end{equation}
where $e(x)=\frac{1}{T_{x}} \sum_{i=1}^{T_{x}}e(x_{i})$ is the embedding of source sentence $x$ which is simply calculated by the average of embeddings of all source tokens. $e(y)$ is the embedding of target sentence $y$ which is defined in the same way. 

As the regularization in Equation~(\ref{equ:align}) just ensures the coarse alignment between the sentence embeddings which is simply the summation of each word embeddings, it misses the fine-grained token-level alignment. Therefore, we propose the word-level adversary learning, considering we do not have the supervision signal of word-level mapping. Specifically, we use Generative Adversarial Network (GAN)~\citep{goodfellow2014generative} to regularize the the projection matrix $W$,
which is widely used in NLP tasks such as unsupervised word translation~\citep{conneau2017word} and text generation~\citep{yu2017seqgan}.
The discriminator $f_D$ takes an embedding as input and outputs a confidence score between $0$ and $1$
to differentiate the embeddings mapped from source tokens, i.e., $E_{z}$, and the ground truth embedding of the target tokens, i.e., $E_{y}$, during training. The linear mapping function $f_{G}$ acts as the generator whose goal is to make $f_{G}$ able to provide plausible $E_{z}$ that is indistinguishable to $E_{y}$ in the embedding space, to fool the discriminator. We implement the discriminator by a two-layers multi-layer perceptron~(MLP). Although other architectures such as CNNs can also be chosen, we find that the simple MLP has achieved fairly good performance.

Formally, given the linear mapping function $f_{G}(\cdot;W)$, i.e., the generator, and the discriminator $f_{D}(\cdot;\theta_{D})$, the adversarial training objective $L_{\textrm{adv}}$ can be written as:
\begin{equation} \label{equ:adv_loss}
L_{\textrm{adv}}(x,y) = \min_{W} \max_{\theta_{D}} V_{\textrm{word}}(f_{G}, f_{D}),
\end{equation}
where $V_{\textrm{word}}$ is the word-level value function which encourages every word in $z$ and $y$ to be distinguishable:
\begin{equation}
\begin{aligned}
\label{equ:v_word}
V_{\textrm{word}}(f_{G}, f_D) =&\mathbb{E}_{e(y_{i}) \sim E_{y}} [\log f_D(e(y_{i}))] +  \\
&\mathbb{E}_{e(x_{j}) \sim E_{x}} [\log(1-f_D(f_{G}(e(x_{j}))))],  
\end{aligned}
\end{equation}
where $e(x_{j})$ and $e(y_{i})$ indicates the embedding of $j$-th source and $i$-th target token respectively. 
In conclusion, for each training pair $(x, y)$, along with the original negative log-likelihood loss $L_{\textrm{neg}}(x,y)$ defined in Equation~(\ref{equ:neg_loss}), the total loss function of our model is:
\begin{equation}
\begin{aligned}
\label{equ:total_loss}
\min_{\Theta} \max_{\theta_{D}} L(x,y) &= L_{\textrm{neg}}(x,y;\theta_{\textrm{enc}}, \theta_{\textrm{dec}}) + \\
&\mu L_{\textrm{align}}(x,y; W) + \lambda L_{\textrm{adv}}(x,y;\theta_{D}, W), 
\end{aligned}
\end{equation}
where $\Theta=(\theta_{\textrm{enc}}, \theta_{\textrm{dec}}, W)$ and $\theta_{D}$ consist of all parameters that need to be learned, while $\mu$ and $\lambda$ are hyper-parameters that control the weight of different losses.

\subsection{Discussion}
The approach of phrase-table lookup is simple and efficient. It achieves considerable performance in experiments by providing direct token-level enhancements, when the phrase table is good enough. However, when training data is messy and noisy, the generated phrase table might be of low quality and consequently hurts NAT model training. We observe that the phrase table trained by Moses can obtain fairly good performance on small and clean datasets such as IWSLT14 but very poor on big and noisy datasets such as WMT14. See Section~\ref{sec:exp_ana} for more details.
In contrast, the approach of embedding mapping learns to adjust the mapping function together with the training of NAT models, resulting in more stable results. 

As for the two components proposed in embedding mapping, the sentence-level alignment $L_{\textrm{align}}$ leverages bilingual supervisions which can well guide the learning of the mapping function,
but lacks the fine-grained word-level mapping signal; word-level adversary loss $L_{\textrm{adv}}$ can provide complimentary information to $L_{\textrm{align}}$. Our ablation study in Section~\ref{sec:exp_ana} (see Table~\ref{tab:ablation_study}) verify the benefit of combining the two loss functions.
\section{Experimental Setup}
\label{sec:exp_set}

\subsection{Datasets}
\label{sec:exp_set_sub}
We evaluate our model on three widely used public machine translation datasets: IWSLT14 De-En\footnote{https://wit3.fbk.eu/}, WMT14 En-De\footnote{https://www.statmt.org/wmt14/translation-task} and WMT16 En-Ro\footnote{https://www.statmt.org/wmt16/translation-task}, 
which has $153$K/$4.5$M/$2.9$M bilingual sentence pairs in corresponding training sets.
For WMT14 tasks, \texttt{newstest2013} and \texttt{newstest2014} are used as the validation and test set respectively. For the WMT16 En-Ro task, \texttt{newsdev2016} is the validation set  and \texttt{newstest2016} is used as the test set. For IWSLT14 De-En, we use 7K data split from the training set as the validation set and use the concatenation of \texttt{dev2010}, \texttt{tst2010}, \texttt{tst2011} and \texttt{tst2012} as the test set, which is widely used in prior works~\citep{ranzato2015sequence,bahdanau2016actor}. All the data are tokenized and segmented into subword tokens using byte-pair encoding~(BPE)~\citep{sennrich2015neural} , and we share the source and target vocabulary and embeddings in each language pair.
The phrase table is extracted from each training set by Moses~\citep{koehn2007moses}, and we follow the default hyper-parameters in the toolkit.

\subsection{Model Configurations}
\label{sec:arch}


We follow the same encoder and decoder architecture as Transformer~\citep{vaswani2017attention}. 
The encoder is composed by multi-head attention modules and feed forward networks , which are all fully parallelizable. In order to make the decoding process parallelizable, we cannot use target tokens as decoder input cause such strong signals are unavailable while inference. Instead, we use the input introduced in the Section~\ref{sec:model}. There exists the problem of length mismatch between the decoder input $z$ and the target sentence, which is solved by a simple and efficient method. 
Given the decoder input candidate $\tilde{z}=(\tilde{z}_{1},...,\tilde{z}_{T_{\tilde{z}}})$ which is either provided by phrase-table lookup or Equation~(\ref{equ:transform_ex}), the $j$-th element of the decoder input $z=(z_{1},...,z_{T_{y}})$ is computed as $z_{j}=\sum_{i}w_{ij} \cdot e(\tilde{z_{i}})$, where $w_{ij}=\exp (-(j-j^{'}(i))^{2}/\tau)$, and $j^{'}(i)=i \cdot \frac{T_{y}}{T_{\tilde{z}}}$, and $\tau$ is a hyper-parameter controlling the sharpness of the function, which is set to $0.3$ in all tasks.

We also use multi-head self attention and encoder-to-decoder attention, as well as feed forward networks for decoder, as used in Transformer~\citep{vaswani2017attention}. Considering the enhanced decoder input is of the same word order of the source sentence, we add the multi-head positional attention to rearrange the local word orders within a sentence, as used in NART~\citep{gu2017non}.
Therefore, the three kinds of attentions along with residual connections~\citep{he2016deep} and layer normalization~\citep{ba2016layer} constitute our model.

To enable a fair comparison, we use same network architectures as in NART~\citep{gu2017non}. Specifically, for WMT14 and WMT16 datasets, we use the default hyper-parameters of the \texttt{base} model described in~\citet{vaswani2017attention},
whose encoder and decoder both have $6$ layers and the size of hidden state and embeddings are set to $512$, and the number of heads is set to $8$.
As IWSLT14 is a smaller dataset, we choose to a smaller architecture as well, which consists of a $5$-layer encoder and a $5$-layer decoder. The size of hidden state and embeddings are set to $256$, and the number of heads is set to $4$.



\subsection{Training and Inference}
We follow the optimizer settings in~\citet{vaswani2017attention}. Models on WMT/IWSLT tasks are trained on $8$/$1$ NVIDIA M40 GPUs respectively. We set $\mu = 0.1$ and $\lambda = 1.0$ in Equation~(\ref{equ:total_loss}) for all tasks to ensure $L_{\textrm{neg}}$, $L_{\textrm{align}}$ and $L_{\textrm{adv}}$ are in the same scale.
We implement our model on Tensorflow~\citep{abadi2016tensorflow}.
We provide detailed description of the knowledge distillation and the inference stage below.

\noindent \textbf{Sequence-Level Knowledge Distillation}~~~During training, we apply the same knowledge distillation method used in~\citep{kim2016sequence,gu2017non,li2019hintbased}. We first train an autoregressive teacher model which has the same architecture as the non-autoregressive student model, and collect the translations of each source sentence in the training set by beam search, which are then used as the ground truth for training the student. By doing so, we provide less noisy and more deterministic training data
which make the NAT model easy to learn~\citep{kim2016sequence,ott2018analyzing,gong2019aaai}.
Specifically, we pre-train the state-of-the-art Transformer~\citep{vaswani2017attention} architecture as the autoregressive teacher model, and the beam size while decoding is set to $4$. 

\noindent \textbf{Inference}~~~While inference, we do not know the target length $T_{y}$. Therefore we first calculate the average ratio between target and source sentence length in the training set which is denoted as $\alpha$, then
predict the target length ranging from $\big[ \lfloor \alpha \cdot T_{\tilde{z_{i}}}-B \rfloor, \lfloor \alpha \cdot T_{\tilde{z_{i}}} + B \rfloor \big]$ where $\lfloor \cdot \rfloor$ denotes the rounding operation. This length prediction method depends on the intuition that the length of source sentence and target sentence is similar, where $B$ is half of the searching window.
$B=0$ indicates the greedy output that only generates a single translation result for a source sentence. While $B \geq 1$, there will be multiple translations for one source sentence, therefore we utilize the autoregressive teacher model to rescore and select the final translation. 
While inference, $\alpha$ is set to $1.1$ for English-to-Others tasks and $0.9$ for Others-to-English tasks, and we try both $B=0$ and $B=4$ which result in $1$ and $9$ candidates. 
We use BLEU scores~\citep{papineni2002bleu} as the evaluation metric\footnote{We report tokenized and case-sensitive BLEU scores for WMT14 En-De and WMT16 En-Ro to keep consistent with NART~\citep{gu2017non}, as well as tokenized and case-insensitive scores for IWSLT14 De-En, which is common practices in literature~\citep{wu2016google,vaswani2017attention}.}.

As for the efficiency, the decoder input $z$ is obtained through table-lookup or the multiplication between dense matrices, which brings negligible additional latency. The teacher model rescoring procedure introduced above is fully parallelizable as it is identical to the teacher forcing training process in autoregressive models, and thus will not increase the latency much. We analyze the inference latency per sentence and demonstrate the efficiency of our model in experiment.
 

\section{Results}

\begin{table*}[htb]
\centering
\begin{tabular}{l|cccc|rr}
\toprule
\multicolumn{1}{c|}{}& \multicolumn{2}{c}{\textbf{WMT14}} & \multicolumn{1}{c}{\textbf{WMT16}} & \multicolumn{1}{c|}{\textbf{IWSLT14}} \\
\textbf{Models}   & \multicolumn{1}{c}{En$-$De} & \multicolumn{1}{c}{De$-$En} & \multicolumn{1}{c}{En$-$Ro} & \multicolumn{1}{c|}{De$-$En} & \multicolumn{2}{c}{Latency / Speedup}  \\
\midrule
LSTM-based S2S~\citep{wu2016google} & $24.60$ & / & / & $28.53^{\dagger}$ & / & /\\
Transformer~\citep{vaswani2017attention}    & $27.41^{\dagger}$ & $31.29^{\dagger}$ & $35.61^{\dagger}$ & $32.55^{\dagger}$ & $607$ ms & $1.00 \times$\\
\midrule
LT~\citep{kaiser2018fast} & $19.80$ & / & / & / & $105$ ms & $5.78 \times$\\
LT (rescoring $10$ candidates)     & $21.00$ & / & / & / & / & / \\
LT (rescoring $100$ candidates)    & $22.50$ & / & / & / & / & / \\
NART~\citep{gu2017non}                 & $17.69$ & $21.47$ & $27.29$ &  $22.95^{\dagger}$ & $39$ ms & $15.6 \times$\\
NART (rescoring $10$ candidates)     & $18.66$ & $22.41$ & $29.02$& $25.05^{\dagger}$ & $79$ ms & $7.68 \times$\\
NART (rescoring $100$ candidates)    & $19.17$ & $23.20$ & $29.79$& / & $257$ ms & $2.36 \times$ \\
IR-NAT~\citep{lee2018deterministic}      & $21.54$ & $25.42$ & $29.66$ & / & $254^{\dagger}$ ms & $2.39 \times$\\
\midrule
Phrase-Table Lookup & $6.03$ & $11.24$ & $9.16$& $15.69$ & / & / \\
\textbf{ENAT Phrase-Table Lookup} & $20.26$ & $23.23$ & $29.85$ & $25.09$ & $25$ ms & $24.3 \times$ \\
\textbf{ENAT Phrase-Table Lookup} (rescoring $9$ candidates) & $23.22$ & $\textbf{26.67}$ & $34.04$& $\textbf{28.60}$ & $50$ ms & $12.1 \times$ \\
\textbf{ENAT Embedding Mapping} & $20.65$ & $23.02$ & $30.08$ & $24.13$ & $\textbf{24}$ ms & $\textbf{25.3} \times$ \\
\textbf{ENAT Embedding Mapping} (rescoring $9$ candidates) & $\textbf{24.28}$ & $26.10$ & $\textbf{34.51}$& $27.30$ & $49$ ms & $12.4 \times$ \\
\bottomrule
\end{tabular}
\caption{BLEU scores on WMT14 En-De, WMT14 De-En, WMT16 En-Ro and IWSLT14 De-En tasks. ``/'' indicates the corresponding result is not reported and ``$\dagger$'' means results are produced by ourselves. We also list the inference latency compared with previous works.
ENAT with rescoring $9$ candidates indicates results when $B=4$, otherwise $B=0$.}
\label{tab:bleu_results}
\end{table*}

\begin{table*}[htb]
\small
\centering
\begin{tabular}{r|l}
\toprule
\multirow{2}{*}{Source:} & hier ist ein foto, das ich am nördlichen ende der baffin-inseln aufnahm, als ich mit inuits auf die narwhal-jagd ging. \\
& und dieser mann, olaya, erzählte mir eine wunderbare geschichte seines großvaters. \\
\hline
\multirow{2}{*}{Target:} & this is a photograph i took at the northern tip of baffin island when i went narwhal hunting with some inuit people, \\
& and this man, olayuk, told me a marvelous story of his grandfather. \\
\hline
\multirow{2}{*}{Teacher:} & here's a photograph i took up at the northern end of the fin islands when i went to the narwhal hunt, \\
& and this man, olaya, told me a wonderful story of his grandfather. \\
\hline
\multirow{2}{*}{NART:} & here's a photograph that i took up the north end of of the baffin fin when i with iuits went to the narwhal hunt,\\
& and this guy guy, ollaya. \& lt; em \& gt; \& lt; / em \& gt; \\
\hline
\multirow{2}{*}{PT:} & so here's a photo which i the northern end the detected when i was sitting on on the went. \\
& and this man , told me a wonderful story his's. \\
\hline
\multirow{2}{*}{ENAT Phrase:} & here's a photograph i took up at the end of the baffin islands i went to the nnarwhal hunting hunt, \\
& and this man, olaaya told me a wonderful story of his grandfather. \\
\hline
\multirow{2}{*}{ENAT Embedding:} & here's a photograph that i took on the north of the end of the baffin islands, when i went to nuits on the narhal hunt, \\ 
&and this man, olaya, told me a wonderful story of his grandfather. \\
\midrule
\midrule
\\[-1em]
Source: & ich freue mich auf die gespräche mit ihnen allen! \\
\hline
\\[-0.5em]
Target: & i look forward to talking with all of you. \\
\hline
\\[-0.5em]
Teacher: & i'm happy to talk to you all! \\
\hline
\\[-0.5em]
NART: & i'm looking to the talking to to you you. \\
\hline
\\[-0.5em]
PT: & i look forward to the conversations with you all! \\
\hline
\\[-0.5em]
ENAT Phrase: & i'm looking forward to the conversations with all of you. \\
\hline
\\[-0.5em]
ENAT Embedding: & i'm looking forward to the conversations to all of you. \\
\bottomrule
\end{tabular}
\caption{Case studies on IWSLT14 De-En task. ENAT Phrase and ENAT Embedding denotes the proposed phrase-table lookup and embedding mapping methods respectively. PT indicates the phrase-table lookup results, which serves as the decoder input to ENAT Phrase method. We collect the results of NART with rescoring $10$ candidates and set $B=4$ while inference for our methods to confirm a fair comparison.}
\label{tab:case_study}
\end{table*}

\begin{table}[htb]
\small
\centering
\begin{tabular}{l|c c}
\toprule
Approach & Decoder Input &  NAT Result \\
\midrule
Word-Table Lookup & $3.54$ & $19.16$ \\
Phrase-Table Lookup & $\textbf{6.03}$ & $\textbf{20.33}$ \\
\bottomrule
\end{tabular}
\caption{The BLEU scores when varying the quality of decoder input on WMT14 En-De task. We set $B=0$ in the inference for the NAT result.}
\label{tab:quality_of_z_wmt}
\end{table}

\subsection{Translation Quality and Inference Latency}
\label{sec:exp_result}

We compare our model with 
non-autoregressive baselines including
NART~\citep{gu2017non},
a semi-non-autoregressive model Latent Transformer~(LT)~\citep{kaiser2018fast} which incorporates an autoregressive module into NART, as well as Iterative Refinement NAT~(IR-NAT)~\citep{lee2018deterministic} which trains extra decoders to iteratively refine the translation output, and we list the ``Adaptive'' results reported in their paper.
We also compare with strong autoregressive baselines that based on LSTM~\citep{wu2016google,bahdanau2016actor} 
and self-attention~\citep{vaswani2017attention}. We also list the translation quality purely by lookup from the phrase table, denoted as Phrase-Table Lookup, which serves as the decoder input in the hard model. For inference latency, the average per-sentence decoding latency on WMT14 En-De task over the \texttt{newstest2014} test set is also reported, which is conducted on a single NVIDIA P100 GPU to keep consistent with NART~\citep{gu2017non}. Results are shown in Table~\ref{tab:bleu_results}.

Among different datasets, our model achieves state-of-the-art performance all non-autoregressive baselines. Specifically, our model outperforms NART with rescoring $10$ candidates from
$4.26$ to $5.62$ BLEU score on different tasks. 
Comparing to autoregressive models, our model
is only $1.1$ BLEU score behind
its Transformer teacher at En-Ro tasks, and we also outperforms the state-of-the-art LSTM-based baseline~\citep{wu2016google} on IWSLT14 De-En task. 
The promising results demonstrate that the proposed method can make the decoder easy to learn by providing a strong input close to target tokens and result in a better model. For inference latency, NART needs to first predict the fertilities of source sentence before the translation process, which is slower than the phrase-table lookup procedure and matrix multiplication in our method. Moreover, our method outperforms NART with rescoring $100$ candidates on all tasks, but with nearly $5$ times faster, which also demonstrate the advantages of the enhanced decoder input.

\begin{figure}[tb]
\centering
\centerline{\includegraphics[width=0.8\columnwidth]{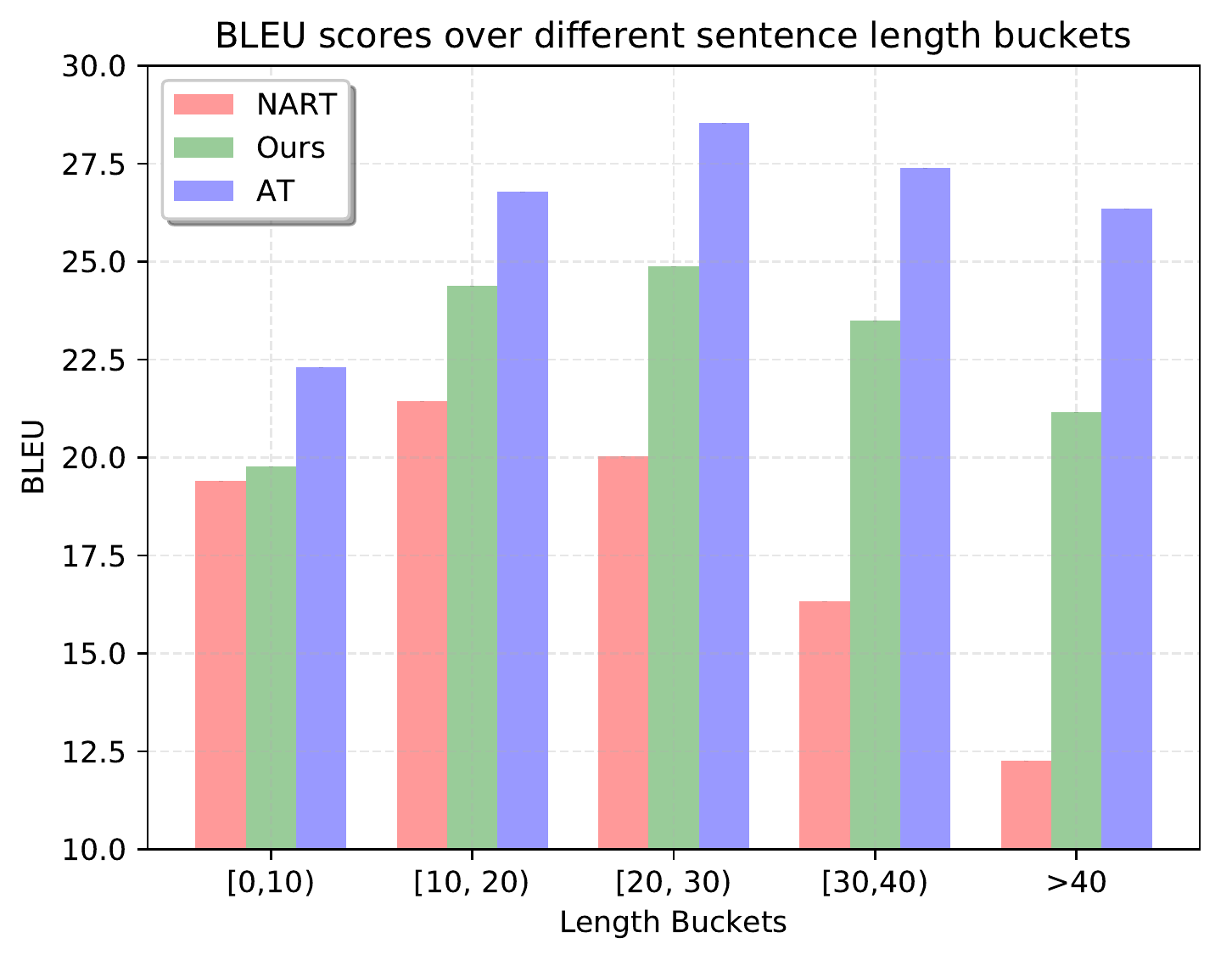}}
\caption{The BLEU scores comparison between AT, NART, and our method over sentences in different length buckets on \texttt{newstest2014}. Best view in color.}
\label{fig:buckets}
\end{figure}

\noindent \textbf{Translation Quality w.r.t Different Lengths}~~~We compare the translation quality between AT~\citep{vaswani2017attention}, NART~\citep{gu2017non} and our method with regard to different sentence lengths. We conduct the analysis on WMT14 En-De test set and divide the sentence pairs into different length buckets according to the length of reference sentence. The results are shown in Figure~\ref{fig:buckets}. It can be seen that as sentence length increases, the accuracy of NART model drops quickly and the gap between AT and NART model also enlarges. Our method achieves more improvements over the longer sentence, which demonstrates that NART perform worse on long sentence, due to the weak decoder input, while our enhanced decoder input provides strong conditional information for the decoder, resulting more accuracy improvements on these sentences.

\subsection{Case Study}
\label{sec:exp_case}
We conduct several case studies on IWSLT14 De-En task to intuitively demonstrate the superiority of our model, listed in Table~\ref{tab:case_study}.

As we claimed in Section~\ref{sec:intro}, the NART model tends to repetitively translate same words or phrases and sometimes misses meaningful words, as well as performs poorly while translating long sentences. In the first case, NART fails to translate a long sentence due to the weak signal provided by the decoder input, while both of our models successfully translate the last half sentence thanks to the strong information carried in our decoder input. As for the second case, NART translates ``to you'' twice, and misses ``all of'', which therefore result in a wrong translation, while our model achieves better translation results again.


\subsection{Method Analysis}
\label{sec:exp_ana}

\noindent \textbf{Phrase-Table Lookup v.s. Embedding Mapping}~~~We have proposed two different approaches to provide decoder input with enhanced quality, and we make a comparison between the two approaches in this subsection.

According to Table~\ref{tab:bleu_results}, the phrase-table lookup achieves better BLEU scores in IWSLT14 De-En and WMT14 De-En task, and the embedding mapping performs better on the other two tasks. We find the performance of the first approach is related to the quality of phrase table, which can be judged by the BLEU score of the Phrase-to-Phrase translation. As IWSLT14 De-En is a cleaner and smaller dataset, the pre-trained phrase table tends to have good quality (with BLEU score 15.69 as shown in Table~\ref{tab:bleu_results}), therefore it is able to provide an accurate enough signal to the decoder. Although WMT14 En-De and WMT16 En-Ro dataset are much larger, the phrase tables are of low quality (with BLEU score 6.03 in WMT14 En-De and 9.16 in WMT16 En-Ro), which may provides noise signals such as missing too much tokens and misguide the learning procedure. Therefore, our embedding mapping outperforms the phrase-table lookup by providing implicit guidance and allow the model adjust the decoder input in a way of end-to-end learning.



\noindent \textbf{Varying the Quality of Decoder Input}~~~We study how the quality of decoder input influence the performance of the NAT model. We mainly analyze in the phrase-table lookup approach as it is easy to change the quality of decoder input with word-table. After obtained the phrase table by Moses from the training data, we further extract the word table from the phrase table following the word alignments. Then we can utilize word-table lookup by the extracted word table as the decoder input $z$, which provides relatively weaker signals compared with the phrase-table lookup. We measure the BLEU score directly between the phrase/word-table lookup and the reference, as well as between the NAT model outputs and the reference in WMT14 En-De test set, listed in Table~\ref{tab:quality_of_z_wmt}. The quality of the word-table lookup is relatively poor compared with the phrase-table lookup. Under this circumstance, the signal provided by the decoder input will be weaker, and thus influence the accuracy of NAT model.

\begin{table}[tb]
\small
\centering
\begin{tabular}{cc|c}
\toprule
$L_{\textrm{align}}$ & $L_{\textrm{adv}}$ & BLEU score \\
\midrule
$\surd$ & $\surd$ & $24.13$ \\
\midrule
 & $\surd$ & $23.53$ \\
$\surd$ &  & $23.74$ \\
\bottomrule
\end{tabular}
\caption{Ablation study of the embedding mapping approach on IWSLT14 De-En task. We set $B=0$ while inference.}
\label{tab:ablation_study}
\end{table}

\noindent \textbf{Ablation Study on Embedding Mapping}~~~We conduct an ablation study in this subsection to study the different components in the embedding mapping approach, i.e., the sentence-level alignment and word-level adversary learning. Results are shown in Table~\ref{tab:ablation_study}. Sentence-level alignment $L_{\textrm{align}}$ slightly outperforms the word-level adversary learning $L_{\textrm{adv}}$. However, adding $L_{\textrm{adv}}$ to $L_{\textrm{align}}$ improves the BLEU score to $24.13$, which illustrates that the complimentary information provided by two loss functions is indispensable.

\section{Conclusion}
We targeted at improving accuracy of non-autoregressive translation models and proposed two methods to enhance the decoder inputs of NAT models: one based on a phrase table and the other one  based on word embeddings. Our methods outperform the baseline on all tasks by BLEU scores ranging from $3.47$ to $5.02$.

In the future, we will extend this study from several aspects. First, we will test our methods on more language pairs and larger scale datasets. Second, we will explore better methods to utilize the phrase table. For example, we may sample multiple candidate target tokens (instead of using the one with largest probability in this work) for each source token and feed all the candidates into the decoder. Third, it is interesting to investigate better methods (beyond the phrase table and word embedding based methods in this work) to enhance the decoder inputs and further improve translation accuracy for NAT models.

\section*{Acknowledgements}
This research was supported by the National Natural Science Foundation of China (No. 61673364, No. 91746301) and the Fundamental Research Funds for the Central Universities (WK2150110008).

\bibliography{nat_with_phrase}
\bibliographystyle{aaai}
\end{document}